%% file: main.tex
\documentclass[11pt]{article}

\usepackage[preprint]{acl}
\usepackage{tcolorbox}
\usepackage{times}
\usepackage{latexsym}
\usepackage{booktabs}

\usepackage[T1]{fontenc}

\usepackage[utf8]{inputenc}

\usepackage{microtype}

\usepackage{inconsolata}
\usepackage{tikz}

\usepackage{graphicx}
\usepackage{rotating}
\usepackage[dvipsnames]{xcolor}
\usepackage[table,xcdraw]{xcolor}

\title{Position: From Noise to  Signal to  Selbstzweck - \\Reframing Human Label Variation in the Era of Post-training in NLP}

\author{Shanshan Xu$^{1,2}$, Santosh T.Y.S.S$^{3}$, Barbara Plank$^{4}$\\
$^{1}$Department of Computer Science, University of Copenhagen, Denmark\\
$^{2}$Faculty of Law, University of Copenhagen, Denmark\\
$^{3}$Amazon
$^{4}$LMU Munich \& Munich Center for Machine Learning (MCML)\\  \texttt{shanshan.xu@di.ku.dk, santoshtyss@gmail.com, b.plank@lmu.de} \\
}

\begin{document}
\maketitle

\begin{abstract}
    \input{text/abstract.tex}
\end{abstract}

\section{Introduction}




The belief in a single ``gold'' human label runs deep in NLP. For decades, annotation pipelines have been designed to minimize inter-annotator disagreement by enforcing a single gold label, treating disagreement as noise to be discarded. Majority vote aggregation, while convenient, risks to flatten the diverse human perspectives. Recently, however, a growing body of work has begun to question this tradition, challenging the assumption that low inter-annotator agreement is necessarily a sign of poor data quality \cite{plank-etal-2014-linguistically,fornaciari2021beyond}. Annotation disagreement is increasingly observed in modern NLP tasks, especially in tasks involving social context. 

Growing work in NLP acknowledges that disagreement is not merely an artifact of error, but often an unavoidable and legitimate \cite{aroyo2015truth,basile-etal-2021-need,prabhakaran-etal-2021-releasing}. This stands in contrast to clear-cut annotation errors, such as misreading task instructions or misclicking a label. To capture this crucial distinction, \citealt{plank2022problem} introduced the notion of human label variation (HLV), urging the community to recognize HLV as a meaningful signal that reflects the pluralistic nature of human values.

\begin{figure*}[t]
    \centering
    \includegraphics[width = 0.93 \textwidth]
    {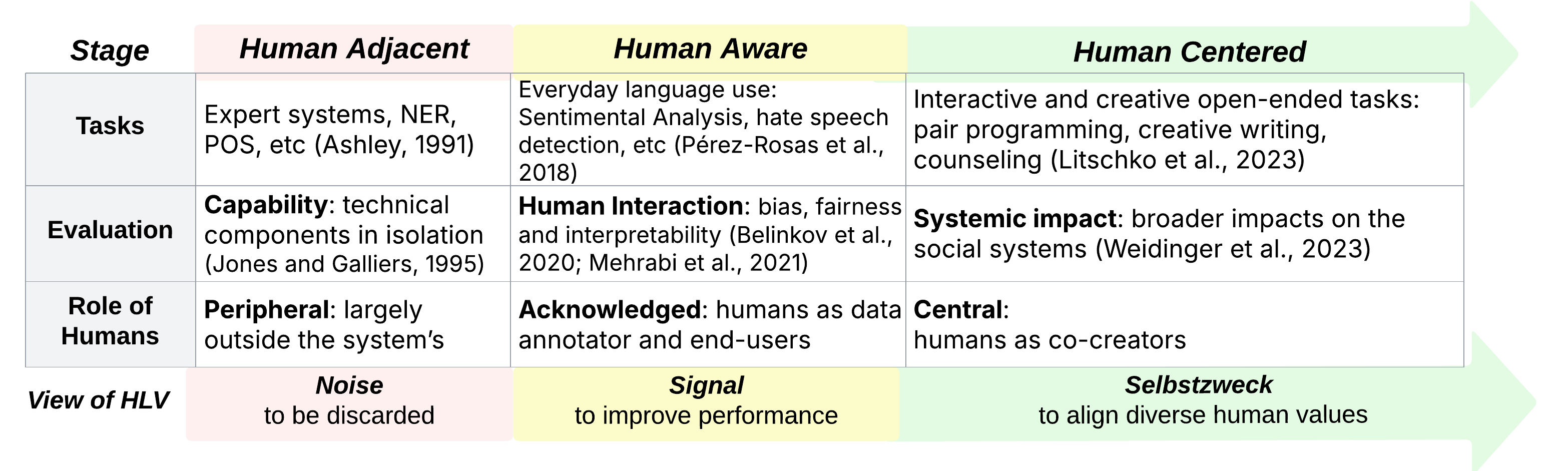}
    \vspace{-0.4cm}
    \footnotesize\caption{Three stages of NLPs evolution toward human-centered systems. The figure illustrates how the changing view of HLV parallels the the  shift in evaluation practices, and growing integration of human pluralism in NLP research. Evaluation row integrates the sociotechnical perspectives of \citealt{weidinger2023sociotechnical}.}
    \label{fig:stages}
    \vspace{-0.5cm}
\end{figure*}

HLV is not only a feature of annotation but also a key channel through which pluralistic social perspectives enter technical systems. Research on AI safety highlights that AI systems are sociotechnical, integrating both model and human components \cite{weidinger2023sociotechnical}. Evaluations of AI must go beyond assessing model capabilities alone and also account for broader social contexts \cite{solaiman2023evaluating,weidinger2024star}. For example in tasks such as hate speech detection, this entails accounting for annotators’ social identities and the societal structures that shape definitions of harm.

Despite growing awareness of pluralistic human alignment \cite{parappan-henao-2025-learning, sorensen-etal-2025-value}, most efforts focus on evaluation frameworks \cite{sorensen2024position} or reward-learning algorithms \cite{chen2406pal}, while the foundation of alignment, the datasets encoding human preferences remains underexamined. \citealt{zhang2025diverging} investigates annotator disagreement in preference dataset (\autoref{fig:hlv_examples} in App \autoref{sec:hlv}), revealing that annotators' diverging preference is systematic rather than random noise, and that aggregation can undermine pluralistic alignment. While their study provides valuable empirical insights, our paper takes a broader conceptual and prescriptive perspective.

 Our contributions are threefold. First, we propose a three-stage conceptual framework (\autoref{fig:stages}) tracing the conceptulization shifts of HLV from \textit{noise} to be eliminated, to \textit{signal} to be modeled and to \textit{Selbstzweck} as intrinsic value. We also explain how these shifts parallel the integration of human pluralism into NLP technologies, as well as the evaluation focus shift in sociotechnical framework\footnote{ \autoref{app:sociotechnical} includes the introduction of the sociotechnical evaluation framework \cite{weidinger2023sociotechnical}.} (\autoref{sec:conceptualization}). Second, we critically examine current preference datasets and identify key limitations that prevent them from capturing genuine human pluralism (\autoref{sec:hlv}). Third, we provide actionable strategies for embedding HLV in future alignment pipelines\footnote{HLV impacts full ML pipeline: data, modeling and evaluation \cite{plank2022problem}. This 4-page position paper focuses on datasets as a key starting point for pluralistic alignment. Further, we focus on post-training data, as it explicitly collects human judgments where annotator-level variation can be directly observed, whereas pretraining corpora consist of passively gathered text.} 
(\autoref{sec:intergrating_hlv}). Rather than presenting new empirical results, we synthesize existing evidence into a coherent framework that reframes how the community should approach human variation in the post-training era.

\section{Human Label Variation}
\label{sec:hlv}
A growing body of work has studied annotation disagreement across NLP tasks, proposing taxonomies to identify its potential sources \cite{uma2021learning, sandri-etal-2023-dont, jiang2022investigating, xu2024through}. Recent work \cite{weber-genzel-etal-2024-varierr} emphasizes the importance of distinguishing \textit{annotation errors} from \textit{HLV}. Annotation errors arise when labels are assigned for invalid reasons. For example, annotator sloppiness, misunderstanding of instructions, or technical issues with the interface. These can often be reduced through better training, clearer guidelines, or improved UI design. By contrast, HLV reflects genuine task ambiguity and human subjectivity, which cannot and should not be eliminated.

As NLP has progressed from well-defined tasks (e.g., POS tagging, NER) toward socially embedded and subjective problems such as emotion categorisation \cite{demszky2020goemotions} and humorousness prediction \cite{simpson2019predicting}, HLV has become increasingly prevalent and unavoidable. Even in natural language inference (NLI), which once considered an “objective” benchmark, researchers observe systematic disagreement (See examples in \autoref{app:hlv_nli}) \cite{pavlick-kwiatkowski-2019-inherent,
jiang2023understanding}.
HLV is also found in preference datasets, as shown in \autoref{fig:hlv_examples} in \autoref{app:example}.  \citet{zhang2025diverging} shows that over 30\% of examples in MultiPref \cite{miranda-etal-2025-hybrid} and HelpSteer2 \cite{wang2024helpsteer} contain disagreement, and that fewer than 25\% of these cases should be attributed to pure annotation error. Nevertheless, major public preference datasets \cite{ouyang2022training, bai2022training} are either annotated by a single annotator or aggregated into a single consensus label. Treating all disagreement as error risks erasing HLV and thereby flattening the pluralism in human values that these datasets could otherwise capture.

\section{Conceptual Evolution of HLV}
\label{sec:conceptualization}
As NLP tasks have grown increasingly subjective, the conceptualisation of HLV has shifted: from being discarded as \textit{noise}, to being recognized as a valuable \textit{signal}. We argue that a further conceptual shift is required in the era of LLMs: HLV, as an embodiment of pluralistic human values, should elevated to \textit{Selbstzweck}, a guiding \textit{design principle} for building genuinely human-centered AI. This reconceptualization reflects both the increasing subjectivity of NLP tasks and the broader re-positioning of human pluralism within NLP system design. 
\citet{Winter2024} propose a framework that categorizes the role of humans in relation to technology, highlighting different levels of integration of human values and agency. Building on this perspective, we introduce a three-stage framework (\autoref{fig:stages}). The trajectory from \textit{Human Adjacent} $\rightarrow$ \textit{Human Aware} $\rightarrow$ \textit{Human Centered} captures how human pluralism have been positioned in NLP over time, moved from the periphery to the core of system design.

\paragraph{Human Adjacent:}
In the early era of NLP (before the 1990s), humans occupied a largely peripheral position in NLP systems. Systems were designed by research labs and corporations primarily to optimize narrowly defined tasks such as text retrieval, syntactic parsing, speech recognition, or named entity recognition. Most systems relied on expert knowledge encoded in “handcrafted” rules or symbolic networks, e.g., 
HYPO \cite{Ashley1991ModelingLA} for legal reasoning in U.S. trade law. The dominant motivation was to increase organizational efficiency and, in the corporate setting, profitability. These tasks were assumed to have a single correct solution, annotated by domain experts. Annotation disagreement was conceptualized as mere error, to be eliminated in order to preserve the “gold standard.” Evaluation focus was on the \textit{capability level}, testing the technical components in isolation, and centering on accuracy and F1 scores against presumed objective truth \cite{jones1995evaluating}. In this stage, humans were \textit{adjacent}: ordinary people largely excluded from shaping how systems were designed.

\paragraph{Human Aware:}
From the 1990s onward, with the rise of the web, deep learning, and later transformer-based models such as BERT \cite{devlin2019bert}, NLP shifted from specialized organizational tools to applications embedded in everyday life, including fake news detection \cite{perez-rosas-etal-2018-automatic} and hate speech detection \cite{sap-etal-2019-risk}. Annotation practices also changed: large-scale crowdsourcing replaced reliance on domain experts, making it possible to scale data collection but also surfacing greater HLV across annotators. At the same time, researchers began to recognize that  NLP applications, especially those used in high-impact domains, can not be tested in isolation without social context.
Models' outputs can affect people’s opportunities or safety in applications like HR screening \cite{li-etal-2020-competence} and legal decision support \cite{xu2023vechr}. Beyond task technical capability, evaulation focus shifted toward \textit{Human-interaction level}, with growing research on interpretability \citep{mehrabi2021survey}, bias and fairness \cite{belinkov2020interpretability}. In this stage, NLP became \textit{human aware}: systems were still largely designed and deployed by organizations, but they increasingly targeted ordinary people as end-users in everyday contexts such as social media.

\paragraph{Human centered:}
Since the launch of ChatGPT \cite{openai2023gpt4}, LLMs have 
transformed NLP from a domain of specialized, compartmentalized tasks \cite{litschko-etal-2023-establishing} into one of open-ended, interactive systems. 
As users increasingly turn to these models for factual information and practical advice, their societal influence has grown rapidly.
Studies demonstrate that LLMs can shape users’ decisions and political beliefs \citep{potter-etal-2024-hidden}. From a sociotechnical perspective, evaluating LLMs requires more than assessing model capabilities or interactions. It demands attention to their \emph{systemic impact} \cite{weidinger2023sociotechnical}: the broader effects on social, economic, and environmental systems, including consequences that appear only at scale, such as political polarization or shifts in public trust.
These developments make the question of alignment more urgent than ever. 
To move toward genuinely \textit{human-centered} AI, pluralism must be treated as a \textit{Selbstzweck}, a guiding design principle. Incorporating HLV proactively in data collection, modelling and evaluation is essential for building systems that are not only aligned but also adaptive to the diversity of human needs, contexts, and worldviews.

                
     



\section{Integrating HLV in Preference Dataset}

\label{sec:intergrating_hlv}

\paragraph{Open Issues}

In the era of LLM, learning from human preferences has emerged as the de facto post-training method for AI Alignment. The purpose of a preference dataset  is to capture human judgments about which model outputs are better, more helpful and harmless. Such judgments are inherently subjective. Annotators may differ in how they weigh factual accuracy, conciseness, empathy, or even stylistic tone, as illustrated in \autoref{fig:hlv_examples}. These differences are not annotation errors but expressions of HLV, reflecting the diversity and pluralism of human values.
The current approach of discarding all the disagreements as mere annotation error\footnote{Distinguishing genuine HLV from annotation errors remains a practical challenge; we provide further discussion in \autoref{app:aed}.} risks over-reducing the inherent HLV which captures the pluralism in human values.

\subsection{Our Suggestions}

\paragraph{Carefully Select the Annotator Pool}


To promote pluralism in human values, managing subjectivity is not about eliminating HLV but about shaping the annotator pool so that it reflects the diversity of the target population. Building preferences datasets is analogous to conducting a survey: a narrow or opaque selection of annotators risks amplifying bias in downstream models \cite{Eckman2024PositionIF}. Yet most LLMs disclose little about annotator recruitment or training; recent analyses \cite{chalkidis2025decoding} show that after 2022, major models virtually stopped reporting  on annotator backgrounds or training practices. To mitigate the selective bias in selecting annotator pool, survey methodology offers guidance, for instance: define the target population, recruit with stratified quotas, and use multi-channel strategies to include underrepresented groups \cite{ahmed2024choose}. Collecting annotator metadata further enables balanced assignments and post-hoc corrections such as post-stratification weighting. For example, \citealt{aroyo2023dices} carefully hired a wide range of annotators to collect diverse opinions of AI safety, providing a basis for analyzing how ratings intersect with demographic categories.


\paragraph{Release Annotator-Level Preferences}
The original HelpSteer2 dataset \cite{wang2024helpsteer} provided only mean scores from five annotators, but its recent update includes the individual annotations. We call for broader adoption of this practice, echoing earlier suggestions on data annotation \cite{prabhakaran-etal-2021-releasing}. Annotator-level labels enable distributional or mixture reward models that capture uncertainty and heterogeneity in preferences rather than collapsing them into a single scalar, leading to more representative policies. \citet{zhang2025diverging} further show that aggregating labels during preference training can harm pluralistic alignment: LLMs are rewarded equally for making decisive choices in both high- and low-agreement cases. This tendency also manifests in LLM-as-Judge evaluation pipelines, which often force a single “winning” response even when preferences diverge. Further, the fine-grained annotations would also allow investigation on the \emph{tied preferences}, where annotators genuinely express no preference between responses (see \autoref{app:tied} for further discussion on HLV vs. tied preferences). Finally, releasing unaggregated annotator-level preferences enhances transparency and enables future research into the sources of disagreement, an essential step toward the pluralistic human alignment.


\paragraph{Leverage Diverse Types of Feedback}
Current preference datasets typically rely on two feedback protocols: relative (pairwise comparisons; \citealt{JMLR:v18:16-634}) and absolute (e.g., Likert-scale ratings; \citealt{10.1145/3357236.3395525}). While relative feedback is simple and widely adopted, it tends to favor assertive or rhetorically salient responses \citep{Kaufmann2023ASO}, obscuring nuanced diverse human judgments. Even with annotator-level labels, binary pairwise comparisons still collapse nuanced diversity in human preferences, whereas absolute feedback better preserve fine-grained perspectives \citep{zhang2025diverging}. Recent studies further demonstrate that combining diverse feedback types, including ratings, comparisons and descriptive evaluations, can produce reward models and downstream policies that surpass single-feedback baselines \citep{metz2025reward}.
To address these limitations, we suggest to
leverage heterogeneous feedback protocols (e.g., scalar ratings, pairwise and n-wise comparisons and demonstrations), to faithfully capture the pluralism of human values that alignment aims to preserve\footnote{Feedback design and alignment algorithms are interconnected. 
App\autoref{sec:appendix} provides ideas for alignment algorithms}.
\paragraph{Pluralism as Reasoning}
In many real-world settings, a single operative decision is unavoidable, for instance, a court must ultimately issue one binding ruling. In such settings, AI systems can preserve pluralism in the reasoning process rather than the final outcome: instead of collapsing disagreement into a single label, systems can document, aggregate, and justify divergent perspectives alongside the unified decision. Legal practice offers a useful analogue: when the judges can not reach a unanimous decision, courts issue binding rulings by majority vote while formally recording dissenting opinions, preserving diverse perspectives for accountability and for future legal reasoning \cite{xu-etal-2024-lens}. Another possible approach is AI as a mediator. Recent work shows an LLM mediator \cite{tessler2024ai} can help small groups to find common ground on divisive political issues by synthesizing participants’ written views into shared group statements.

\section{Conclusion}

In this paper, we trace the evolving view of Human Label Variation (HLV) from \textit{noise}, to \textit{signal}, to \textit{Selbstzweck}; and illustrate how this shift parallels the broader integration of human pluralism into NLP technologies. To advance toward genuinely human-centered AI, we argue for the proactive inclusion of HLV in the creation of preference datasets. While our discussion centers on preference data due to scope limit, the implications of HLV extend across the entire machine learning pipeline, from data collection, to model training, and evaluation. We hope this survey raises awareness within the community and inspires further work on modeling and evaluating alignment in ways that account for pluralistic human values.

\section*{Limitations}

This position paper aims to be concise while synthesizing a broad conceptual shift: the evolving view of Human Label Variation (HLV) mirrors the broader integration of human pluralism into NLP technologies. As such, several limitations should be acknowledged:

First, we acknowledge that incorporating pluralism introduces additional computational and financial costs. Our goal is not to prescribe a specific implementation strategy, but to emphasize that HLV and pluralistic alignment should be treated as first-order design considerations. The three-stage evolution we outline (Human Adjacent $\rightarrow$ Human Aware $\rightarrow$ Human Centered) reflects precisely how technological and societal developments have made it both feasible and necessary to evaluate models beyond pure task performance, even at additional cost.

Second, our suggestions on how to including HLV in dataset creation focuses on preference dataset. Limited by the scope, we do not address the \textit{pretraining} corpora, which likely exerts an even stronger influence on models’ ideological tendencies and value alignment \cite{xu2025better,ceron2025political}. The Selbstzweck principle may generalize to pretraining data collection, e.g., through normative considerations in data filtering or multilingual corpus construction. We focus on post-training datasets, as post-training data like RLHF explicitly collects human judgments where annotator-level variation can be directly observed, whereas pretraining corpora consist of passively gathered text.

Third, our analysis does not comprehensively cover the whole pipeline of alignment techniques. As pointed out by \citealt{plank2022problem}, implications of HLV span the entire machine learning pipeline, from data collection and model training to evaluation. 
While we have addressed on the preference dataset collection in \autoref{sec:intergrating_hlv} and briefly discussed on alignment algorithms in \autoref{sec:appendix}, future work could explore new evaluation metrics, for example by developing metrics such as:

\begin{itemize}
\item \textbf{Pluralism fidelity} assesses how closely the model’s conditional output distribution matches the empirical human response distribution for a given context. This is related to distributional pluralism in recent work \cite{sorensen2023value} on pluralistic alignment roadmap. In practice, pluralism fidelity can be quantified using standard divergence measures such as  KL divergence, or Wasserstein distance as used in prior work \cite{durmus2023towards,santurkar2023whose}.

\item \textbf{Minority regret} quantifies the utility loss (regret) for minority preference groups relative to an optimal group-aligned model, as defined by in \citealt{raghavan2018externalities}. This mirrors recent work such as MaxMin-RLHF \cite{chakraborty2024maxmin}, which optimizes for the worst‑served group rather than the population average.
\end{itemize}

Finally, given the rapid pace of progress in this area, some very recent developments may not be included. As a position paper, our aim is to provoke discussion and outline future research directions, rather than to offer comprehensive solutions or empirical evaluations. We encourage further work that operationalizes these principles in a broader range of pluralistic, linguistic, and technological settings. We hope this paper contributes to raising awareness within the community and encourages further exploration of methods for collection, modeling and evaluating human pluralism into NLP research and practice.

\section*{Information about use of AI assistants}
In the preparation of this work, the authors utilized ChatGPT and Gemini to polish the writing and improve the coherence of the manuscript. Before the submission, the authors conducted a thorough review and made necessary edits to the content, taking full responsibility for the final version of the text.




\bibliography{main}
\newpage
\appendix


\label{sec:appendix}

\input{text/appendix}

\end{document}

%% file: text/abstract.tex

Human Label Variation (HLV) refers to legitimate disagreement in annotation that reflects the diversity of human perspectives rather than mere error. Long treated in NLP as \textit{noise} to be eliminated, HLV has only recently been reframed as a \textit{signal} for improving model robustness. With the rise of large language models (LLMs) and post-training methods such as human feedback-based alignment, the role of HLV has become increasingly consequential. Yet current preference-learning datasets routinely collapse multiple annotations into a single label, flattening diverse perspectives into artificial consensus. Preserving HLV is necessary not only for pluralistic alignment but also for sociotechnical safety evaluation, where model behavior must be assessed in relation to human interaction and societal context.
This position paper argues that preserving HLV as an embodiment of human pluralism must be treated as a \textit{Selbstzweck}, an intrinsic value in itself. We analyze the limitations of existing preference datasets and propose actionable strategies for incorporating HLV into dataset construction to better preserve pluralistic human values.

%% file: text/appendix.tex
\section{Sociotechnical Evaluation Framework}
\label{app:sociotechnical}
\citealt{weidinger2023sociotechnical} proposes a three-layer sociotechnical framework for evaluating safety risks of generative AI systems, motivated by the idea that generative AI systems are intertwined with human and model. 
Context determines whether a capability causes harm, so evaluation must consider context beyond model-only tests. The framework includes the following three layers:

\paragraph{Layer 1: Capability evaluation}: focuses on testing the technical components of AI systems in isolation (e.g., outputs, model behavior, training data,  and other artifacts), providing early indicators of potential downstream harms but not sufficient evidence of real-world harm by itself.

\paragraph{Layer 2: Human interaction evaluation}: evaluates what happens when people actually interact with the system, emphasizing that safety depends on the context: who uses the system, for what purpose, and in what setting.

\paragraph{Layer 3: Systemic impact evaluation}: assesses broader impacts on the systems in which AI is embedded (society, institutions, economy, environment), including effects that emerge only at scale (e.g.,political polarisation or changes to trust in public media).

\section{HLV examples from the NLI dataset}
\label{app:hlv_nli}
 HLV has become increasingly prevalent and unavoidable. Even in natural language inference (NLI), which once considered an “objective” benchmark, researchers observe systematic disagreement \cite{pavlick-kwiatkowski-2019-inherent,
jiang2023understanding}.
 For example, given the \textit{Premise}: \textit{“Technological advances generally come in waves that crest and eventually subside”} and the \textit{Hypothesis}: \textit{“Advances in electronics come in waves”}, 82 annotators labeled the pair as \textit{Entailment}, 17 as \textit{Neutral}, and 1 as \textit{Contradiction} \cite{jiang2022investigating,nie-etal-2020-learn}. The disagreement arises from annotators' different interpretations of the lexical relationship between “electronics” and “technological advances.”

\section{HLV examples from the preference datasets}
\label{app:example}
\autoref{fig:hlv_examples} shows two \textbf{HLV}examples from the preference dataset MutliPref \cite{miranda-etal-2025-hybrid}

\input{figs/hlv_pref.tex}

\section{HLV vs. Tied Preferences}
\label{app:tied}
We regard HLV and tied preferences conceptually distinct. HLV refers to legitimate difference across annotators, whereas tied preferences reflect legitimate indifference or equal valuation between options. If multiple annotators select “tie,” this indicates shared indifference rather than HLV/disagreement.

Importantly, we believe that ties are not noise. At a deeper level, ties reflect the fact that human values are not always strictly ordered. In real moral and evaluative cognition, some options are incomparable, some are roughly equivalent, and some are lexically ordered. Modeling preferences only as total orders obscures this structure. Representing comparisons as {X > Y, Y > X, Tie} preserves information about both direction and strength. Tie-heavy regions may indicate high reward overlap or preference-insensitive cases, while polarized disagreement signals value-sensitive trade-offs. Recent work that explicitly models ties show that incorporating ties can improve preference learning, e.g. DPO extensions based on Rao–Kupper/Davidson and approaches using ordinal feedback \cite{chen2024extending, liu2024reward}.

\section{HLV vs. Annotation Error}
\label{app:aed}
As discussed in Sec 2, earlier works \cite{xu-etal-2023-dissonance, jiang2022investigating} have attempted to identify disagreement/hlv into different categorizations:  e.g. human subjectivity (HLV) or sloppiness of the annotators (noise). However, reliably separating these sources of variation remains challenging and is still underexplored in current automatic error detection (AED) research.
Action. Recent work \cite{weber-genzel-etal-2024-varierr} uses LLMs to uncover annotation errors versus human label variation. They found that annotator rationales can help contextualize disagreement, but does not fully resolve the identification problem.

\section{Feedback Design and Alignment Algorithm}

Feedback design and alignment algorithms are deeply intertwined: preserving Human Label Variation (HLV) requires alignment methods that explicitly optimize for pluralism. In practice, this means bridging heterogeneous feedback types and modeling disagreement \textit{distributionally}, rather than collapsing it into a single scalar objective.

A growing body of work explores \textit{multi-objective RLHF} (MORLHF; \citealt{xiong2025projection}), which aims to align models with diverse human preferences by interpolating across multiple objectives. Two main families of approaches exist: 
(a) \textbf{Inference-time interpolation}, where models trained on distinct objectives are combined through weight merging at inference \citep{rame2023rewarded,jang2023personalized}; and 
(b) \textbf{Training-time scalarization}, where multiple reward functions are jointly optimized via linear combination \citep{GhaneKanafi2015ANS}.

However, these methods often obscure minority preferences or assume static, universal objectives \citep{rame2023rewarded,GhaneKanafi2015ANS}. To move beyond these limitations, future alignment algorithms should:
\begin{enumerate}
    \item \textbf{Leverage heterogeneous feedback protocols} simultaneously (e.g., scalar ratings, pairwise and n-wise comparisons, demonstrations, corrections);
    \item \textbf{Model HLV distributionally}, rather than collapsing it through scalarization;
    \item \textbf{Enable conditional personalization}, adapting to subgroup or individual value profiles while safeguarding minority perspectives; and
    \item \textbf{Support continual or online learning}, allowing alignment to evolve alongside changing human norms and preferences.
\end{enumerate}

Alignment should not converge toward a single ``best'' policy but instead define a \textbf{navigable policy space} that reflects the distribution of human disagreement. Ultimately, alignment algorithms should make value conflict \textit{transparent and governable}, enabling models to mediate between competing perspectives rather than enforcing false consensus.

%% file: figs/hlv_pref.tex
\begin{figure}[h]
\centering
\footnotesize

\begin{tcolorbox}[
  colback=teal!4,
  left=3pt,right=3pt,
  colframe=teal!70!white,
  title=\textbf{
  Content Preference HLV},
  boxsep=0pt,
  boxrule=0.8pt
]
\footnotesize

{\textbf{Prompt}} : Would embodiment of an LLM be a prerequisite for sentience? \\[1pt]
\textbf{Response \color{BrickRed}A:} No, embodiment of an LLM would not be a prerequisite for sentience~\dots \\[1pt]
\textbf{Response \color{MidnightBlue}B:} The relationship between embodiment and sentience in LLMs remains an open question~\dots \\[1pt]
\textbf{Annotator 1}: Prefers \textbf{\color{MidnightBlue}B} \quad
\textbf{Annotator 2} : Prefers \textbf{\color{BrickRed}A}
\end{tcolorbox}
\vspace{-0.1cm}
\begin{tcolorbox}[
  colback=orange!4,
  colframe=orange!70!white,
  title=\textbf{
  Stylistic Preference HLV},
  rounded corners,
boxsep=0pt,
left=3pt,right=3pt,
  boxrule=0.8pt
]
\footnotesize

{\textbf{Prompt:} Give me recipes for five blue-colored cocktails. \\[1pt]
\textbf{Resp. \color{BrickRed}A:} Blue Hawaiian, Ingredients: 2 oz blue curaçao~\dots \\[1pt]
\textbf{Resp. \color{MidnightBlue}B:} Creating blue cocktails is a fun way to add a vibrant touch ~\dots Here are five blue cocktail recipes~\dots \\[1pt]

\textbf{Annotator 1}: Prefers \textbf{\color{BrickRed}A} \quad

\textbf{Annotator 2}: Prefers \textbf{\color{MidnightBlue}B}
}

\end{tcolorbox}
\vspace{-0.4cm} 
\footnotesize\caption{Two \textbf{HLV}examples from the preference dataset MutliPref \cite{miranda-etal-2025-hybrid}.
The upper one reflects differences in \textit{content} preference, while the lower reflects differences in \textit{ stylistic} preference.}
\label{fig:hlv_examples}
\end{figure}